\documentclass[lettersize,journal]{IEEEtran}
\usepackage{amsmath,amsfonts}
\usepackage{algorithmic}
\usepackage{array}
\usepackage[caption=false,font=normalsize,labelfont=sf,textfont=sf]{subfig}
\usepackage{textcomp}
\usepackage{xurl}
\usepackage{verbatim}
\usepackage{graphicx}
\usepackage{balance}

\usepackage{pdfpages}
\usepackage{float}
\usepackage{subcaption}
\usepackage{hyperref}
\usepackage{enumitem}
\usepackage{multirow}

\begin{document}
\title{Breaking Down the Barriers: Investigating Non-Expert User Experiences in Robotic Teleoperation in UK and Japan}

\author{Florent P Audonnet$^{1,2}$, Andrew Hamilton$^{1}$, Yukiyasu Domae$^{2}$, Ixchel G. Ramirez-Alpizar$^{2}$\\ and Gerardo Aragon-Camarasa$^{1}$
\thanks{This research has been supported by EPSRC DTA No. 2605103 and RSE Saphire scheme (grant no. 3737)}
\thanks{$^{1}$ School of Computing Science, University of Glasgow, G12 8QQ, UK, United Kingdom {\tt\small f.audonnet.1@research.gla.ac.uk; gerardo.aragoncamarasa@glasgow.ac.uk}}%
\thanks{$^{2}$ National Institute of Advanced Industrial Science and Technology}
}

\maketitle
\let\oldUrl\url
\renewcommand{\url}[1]{\href{#1}{#1}}

\begin{abstract}
Robots are being created each year with the goal of integrating them into our daily lives. As such, there is an interest in research in evaluating the trust of humans toward robots. In addition, teleoperating robotic arms can be challenging for non-experts. To reduce the strain put on the user, we created TELESIM, a modular and plug-and-play framework that enables direct teleoperation of any robotic arm using a digital twin as the interface between users and the robotic system. We evaluated our framework using a user survey with three robots and control methods and recorded the user's workload and performance at completing a tower stacking task. However, an analysis of the strain on the user and their ability to trust robots was omitted. This paper addresses these omissions by presenting the additional results of our user survey of 37 participants carried out in United Kingdom. In addition, we present the results of an additional user survey, under similar conditions performed in Japan, with the goal of addressing the limitations of our previous approach, by interfacing a VR controller with a UR5e. Our experimental results show that the UR5e has more towers built. Additionally, the UR5e gives the least amount of cognitive stress, while the combination of Senseglove and UR3 provides the user with the highest physical strain and causes the user to feel more frustrated. Finally, the Japanese participants seem more trusting of robots than the British participants.

\end{abstract}

\section{Introduction}\label{sec:intro}
The advent of Industry 4.0 has fundamentally transformed the manufacturing landscape which has marked a transition from conventional programmable robots to sophisticated, data-driven systems~\cite{shiIndustry40,MEINDL2021120784}. This paradigm shift, coupled with advancements in digital twin technology, has enabled the development of virtual replicas of smart factories to enhance operational efficiency and decision-making processes. As we progress into the era of Industry 5.0, the focus is now on fostering an ecologically and socially responsible industry that prioritizes human-centric values~\cite{adel2022future}. This evolution needs the emergence of machines that extend beyond mere digital replicas. That is, these machines have to evolve into collaborative partners that support humans in navigating complex tasks.

Central to the framework of Industry 5.0 is teleoperation systems, which facilitate remote human-machine collaboration~\cite{adel2022future}. However, extended teleoperation sessions present challenges, notably leading to physical and mental fatigue among operators~\cite{pettinger_reducing_2020}. Moreover, the requisite level of expertise for effective teleoperation still needs to be more adequately defined~\cite{audonnet_telesim_2024}. This highlights the need for further research to establish standardized protocols and training programs to enhance teleoperation practitioners' efficiency and well-being. Therefore, This paper investigates the influence of robotic hardware and participant demographics on performance metrics and user workload in the context of direct teleoperation.

Our previous work introduced TELESIM, a framework for intuitive robotic arm teleoperation~\cite{audonnet_telesim_2024}. While our initial study with over 30 participants demonstrated that non-experts could effectively teleoperate robotic arms, it did not fully address the physical and cognitive strains experienced by users. This critical aspect warrants further investigation to ensure user safety and well-being in teleoperation scenarios~\cite{pettinger_reducing_2020}.

\begin{figure}[t]
    \centering
    \includegraphics[width=0.95\linewidth]{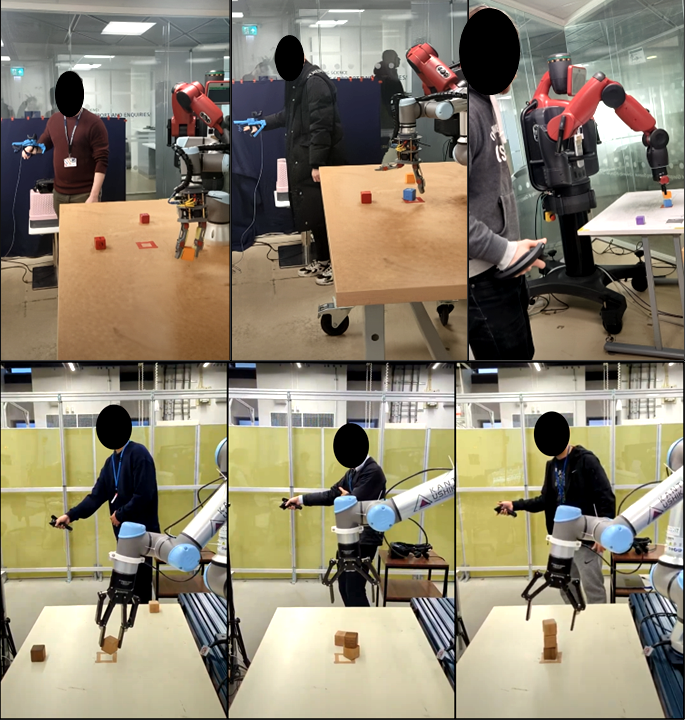}
    \caption{Our framework TELESIM is being used to control a UR3 Robot (top-right, top-center) and a Baxter Robot (top-left) in the United Kingdom and a UR5e robot in Japan (bottom)}
    \label{fig:overview}
\end{figure}

To address these gaps and expand on our previous findings, this paper rigorously explores teleoperation's impact on users' mental and physical health and their relationship with robots during task performance. For this, we conducted a large-scale, international user survey across Japan and the United Kingdom, involving over 70 participants from diverse backgrounds. The study used three robots with varying ranges and speeds: the Universal Robot 3, Universal Robot 5e, and Rethink Robotics Baxter. Participants performed a standardized 3-cube tower stacking task for 10 minutes, which allowed us to compare robot types and user demographics directly. Our contributions are thus the following:
\begin{enumerate}
    \item A large-scale, cross-cultural comparison of teleoperation performance and user experience between Japan and the United Kingdom (UK), involving 74 participants from diverse backgrounds.
    \item An in-depth analysis of the relationship between robot capabilities (e.g., reach, stability, control method) and teleoperation performance derived from experiments with three robotic arms: Baxter, UR3, and UR5e.
    \item A comprehensive evaluation of user workload during teleoperation using the NASA-TLX questionnaire. We found that the difference in hardware impacts the mental workload and that the user's frustration is not linked to the performance of the teleoperation.
    \item An investigation into the differences in trust towards robots between the Japanese and the British participants, challenging previous findings that the British users trusted robots more than the Japanese participants~\cite{bartneck_influence_2007}.
\end{enumerate}

The rest of this paper is organized as follows. In Section \ref{sec:background}, we explore the current state of the art and the research gap. Specifically, in Section \ref{sec:teleop}, we review existing teleoperation frameworks and the methodologies other researchers have employed to evaluate them. Then, in Section \ref{sec:teleop_eval}, we explore ways the user's workload during teleoperation has been assessed. Finally, in Section \ref{sec:background_trust}, we inspect how trust has been recorded and evaluated in teleoperation.
Afterwards, in Section \ref{sec:telesim}, we present a recapitulation of the TELESIM framework and our experimental design implemented in the United Kingdom. Thus, we detail in Section \ref{sec:experiments} a detailed description of our experimental setup, emphasizing the distinctions between our experiments conducted in Japan and the United Kingdom.
Finally, in Section \ref{sec:eval}, we provide a comprehensive analysis and comparison of the workload and performance of our teleoperation framework.

\section{Background}\label{sec:background}

\subsection{Teleoperation Systems}\label{sec:teleop}

Direct teleoperation is recognized as an essential precursor to shared autonomy~\cite{zhang_haptic_2021}. This recognition arises from the considerable cognitive demands placed on users during direct teleoperation~\cite{pettinger_reducing_2020} and the challenges users face in executing precise, millimetre-scale adjustments to a robot's end effector. In medical applications, user movements are often scaled down to enhance precision~\cite{lanfranco_robotic_2004, rakita_motion_2017}. However, such scaling is only universally applicable across some manipulation tasks, particularly those requiring extensive arm movements for object relocation.

While substantial research has been done in the field, only our previous work \cite{audonnet_telesim_2024} has evaluated direct teleoperation across multiple robots without incorporating shared autonomy. Existing literature primarily focuses on shared autonomy to alleviate cognitive strain while maximizing precision. For example, researchers have explored various control methodologies, such as low degree-of-freedom interfaces such as keyboards~\cite{katyal_approaches_2014}, joysticks~\cite{aronson_eye-hand_2018, scherzinger_learning_2023}, touchscreens~\cite{toh_dexterous_2012}, and gamepads~\cite{micire_design_2011}, which have shown improved control levels and reduced mental strain~\cite{katyal_approaches_2014}. With the advent of virtual reality (VR) and augmented reality (AR) technologies, VR controllers~\cite{wang_intent_2021, dafarra_icub3_2022} and smartphones~\cite{mandlekar_roboturk_2018} have been investigated for their applicability in teleoperation. Some studies have also examined motion mapping of the user's body~\cite{chen_intuitive_2012, rosen_communicating_2019} or gaze control~\cite{admon_predicting_2016}, though these methods present challenges such as kinematic discrepancies and increased cognitive load~\cite{pettinger_reducing_2020}. While extensive research has been conducted, the diversity of findings allows for tailored control methods that can be selected based on specific cases. The control method, therefore, remains a choice made on a case-by-case basis.

To address the limitations of direct teleoperation, researchers have emphasized the benefits of shared autonomy in enhancing task success rates. However, the absence of literature focusing solely on direct teleoperation suggests that these limitations may be attributed to the specific implementation of teleoperation in each study.
Notably, researchers have employed either MoveIt! ~\cite{schuppstuhl_annals_2022}, which has its limitations, described in Section \ref{sec:framework}, or custom planning interfaces for robot control, which is either not detailed how teleoperation was implemented~\cite{lin_shared_2020}, or whether teleoperation was developed for the specific shared autonomy framework\cite{javdani_shared_2018, gottardi_shared_2022}. In addition, these studies have predominantly focused on single robotic systems and conducted experiments with small user groups (between 8 and 23 participants), emphasizing different autonomy levels rather than control methods.

\begin{figure*}[!th]
\centering
\includegraphics[width=0.95\textwidth]{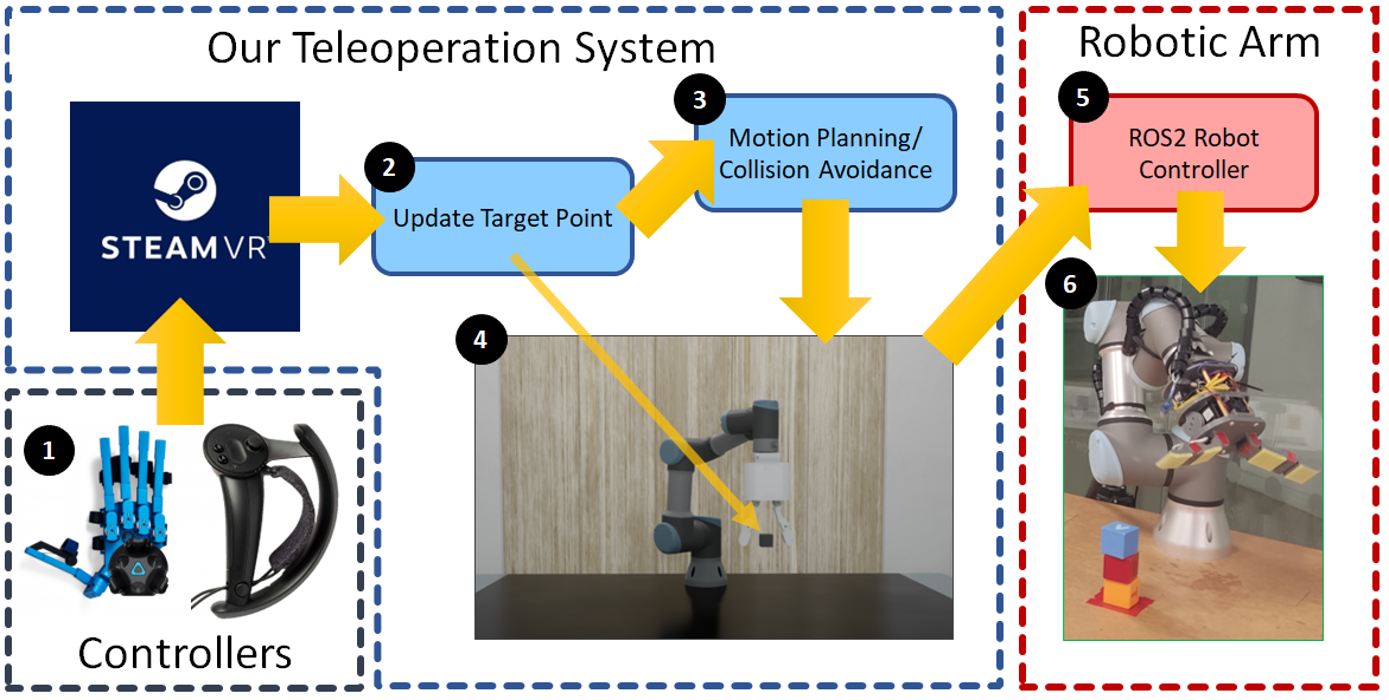}
\caption{Overview of TELESIM. The controllers (in the black dotted line) (1) can be any system that outputs a 3D pose. TELESIM is depicted in the blue dotted line, which accepts the pose given by (1) to update the 3D pose of a cube in the digital twin. The robot then calculates a path to this cube in real-time while avoiding collision with the world (4). Finally, as shown in the red dotted line, TELESIM can be plugged into any robotic system (6) via a ROS2 robot controller (5).}
\label{fig:schema}
\end{figure*}

\subsection{Evaluating Workload for Teleoperation Systems}\label{sec:teleop_eval}

Shared autonomy can potentially enhance direct teleoperation performance, but human factors significantly influence the overall success of tasks. Therefore, an efficient teleoperation system must minimize both physical and cognitive stress on the operator, and assessing user workload is crucial, with the NASA Task Load Index (NASA-TLX)~\cite{hart_nasa-task_2006} being the predominant tool utilized in the literature for this evaluation. Sandra Hart developed the NASA-TLX questionnaire at the NASA Ames Research Center in 1988~\cite{hart_development_1988}, a widely used subjective assessment tool designed to measure perceived workload across various tasks and environments. It evaluates workload based on six dimensions: Mental Demand, Physical Demand, Temporal Demand, Performance, Effort, and Frustration. Two versions of the NASA-TLX are available. The original version includes six 21-point scales for each dimension and a pairwise comparison of each dimension, whereas the "Raw" version omits the pairwise comparison~\cite{hart_nasa-task_2006}. The "Raw" version is often favoured because it is easier to carry out and analyse~\cite{hart_nasa-task_2006}.

In robotic teleoperation, the NASA-TLX is the preferred instrument for assessing workload~\cite{lin_shared_2020, parsa_impact_2022, gottardi_shared_2022}. To our knowledge, only a few research works have opted not to use the NASA or "Raw" NASA questionnaire for robotic teleoperation. That is, some researchers instead decided to use the NASA questionnaire to measure the user's workload using extra devices or metrics. For example, Moya \textit{et al.}~\cite{moya_workload_2017} utilized EEG but corroborated these results with the NASA questionnaire. 
Additionally, researchers such as Naughton \textit{et al.}~\cite{naughton_integrating_2024} were interested solely in the overall NASA score, which they obtained by summing all dimensions. However, it is not the only way to use the NASA questionnaire. Parsa \textit{et al.}~\cite{parsa_impact_2022} conducted an ANOVA analysis on each NASA dimension rather than using the questionnaire score to extract more value from the questionnaire. Others, such as Lin \textit{et al.}~\cite{lin_shared_2020}, focused on specific metrics to emphasize particular results~\cite{torielli_wearable_2024}.
Finally, Bechtel \textit{et al.}~\cite{bechtel_toward_2023} employed a single question for users to self-report their overall workload on a Likert-Scale, similar to the Single Ease Question, which is "Overall, This task was ?" 

\subsection{Trust in Robots}\label{sec:background_trust}

Trust in robots is a critical factor, especially given the recent increase in robots designed to interact with the general population~\cite{bogue_first_2022}. Consequently, researchers have developed various questionnaires tailored to different use cases, as evaluating trust is complex and often requires multiple scales~\cite{wachowiak_survey_2023}. Prominent scales include the General Attitudes Towards Robots Scale (GAToRS)~\cite{koverola_general_2022}, the Robotic Social Attributes Scale (RoSAS)~\cite{carpinella_robotic_2017}, and the Negative Attitude Towards Robots Scale (NARS)~\cite{syrdal_negative_2009}. These scales evaluate social robots with which users can interact physically or verbally or establish an anthropomorphic connection. NARS~\cite{syrdal_negative_2009} remains popular due to its three subscales, which can analyse a broad range of attitudes toward robots: negative attitudes toward interaction situations with robots (S1), negative attitudes toward the social influence of robots (S2), and negative attitudes toward emotions in interaction with robots (S3). Each scale can be employed individually to examine specific cases. NARS is the only trust scale used to evaluate trust in robotic arm manipulation~\cite{tsui_using_2010}. 

\section{Teleoperation Framework}\label{sec:telesim}\label{sec:framework}

TELESIM ~\cite{audonnet_telesim_2024} (Fig. \ref{fig:schema}) is a framework that enables teleoperation of various robotic arms. The framework comprises three components, with the controller (1) being any control system that outputs a 3D pose (described in Sec. \ref{sec:controllers}. TELESIM, (2) in Figure \ref{fig:schema} and described in Sec. \ref{sec:telesimdetails}, computes a real-time motion plan for the pose provided by the controller (1) using Nvidia Isaac Sim ~\cite{noauthor_isaac_2019}, an RTX simulator serving as a digital twin of the environment. The joint states of the robot are then sent through ROS2 to any robotic arm that supports ROS2 control (3) in Fig. \ref{fig:schema}.

\subsection{Controllers}\label{sec:controllers}

Our choice of control method was informed by the work of Gottardi et al.~\cite{gottardi_shared_2022}, who explored integrating multiple control systems, including the combination of VR controllers with tracking bands to monitor user movements, and Rakita et al.~\cite{rakita_motion_2017} who compared several control methods and integrated them into a custom inverse kinematics solver to align the end-effector pose with the user's input. Their findings indicated a preference for VR controllers in completing pick-and-place tasks. Thus, the controller for TELESIM, (2) in Fig. \ref{fig:schema}, consisted of a Steam Index VR controller used to control a Rethink Robotics Baxter robot and the UR5e and a Senseglove to control the UR3. We used the Senseglove for mapping individual finger motions, particularly the index and thumb, to the two fingers of the Yale T42 gripper, described in Section \ref{sec:gripper}. A Vive VR Tracker, mounted on top of the Senseglove, was employed to map user motions into the UR3 robot. 

The Senseglove functions analogously to a glove affixed to the palm and fingertips. For our user survey, we opted for the Senseglove development kit (shown in Figure \ref{fig:schema}(1)) due to its ease of application and adjustment. Moreover, we limited the strapping to the index finger and thumb, corresponding to our two-fingered gripper design. Although the Senseglove SDK is compatible with multiple platforms, integration into our framework required us to port it to ROS2. This was accomplished through the creation of a ROS2 control plugin\footnote{\url{https://github.com/09ubberboy90/senseglove\_ros2\_ws}}. This adaptation enabled the seamless incorporation of the Senseglove functionality within our TELESIM framework.
Using this plugin, we are able to read the state of the user's hand by aggregating the joint angles of each finger and scaling them to a range corresponding to the gripper finger being fully opened and fully closed. This scaling aligns with the motor control range, derived from average values recorded in practice for a fully opened hand and a closed hand. This methodology enables a standardized mapping of finger positions to motor control inputs, facilitating the accurate translation of user hand movements to gripper actions.

We use the Python interface for SteamVR to develop an application that continuously monitors the controller's position, subsequently transmitting this data to the rest of the framework via multiple ROS2 (Robot Operating System)~\cite{macenski_robot_2022} publishers, one for each controller. An additional publisher conveys information about button states, facilitating custom integration for potential future research by allowing modifications to their interactions with our TELESIM framework. This design also allows us to incorporate new functionalities, as shown in~\cite{audonnet_immertwin_2024}.

\subsection{TELESIM}\label{sec:telesimdetails}

TELESIM's core functionality is predicated on acquiring a three-dimensional pose from the controller (Fig. \ref{fig:schema}) and subsequently generating a motion plan for any robotic arm. The system's architecture is built upon NVIDIA Isaac Sim and ROS2, as described below.

The robotic systems are dynamically instantiated in Isaac Sim from their Universal Robot Descriptor Files (URDF) to enhance flexibility and adaptability to any robot and make TELESIM a modular and plug-and-play framework. Isaac Sim offers a diverse array of methodologies for motion planning and collision avoidance. For instance, Rapidly Exploring Random Tree (RRT) algorithms~\cite{lavalle_rapidly-exploring_1998}, specifically Nvidia's Lula RRT implementation, offer global solutions primarily suited for static environments. Conversely, Moveit2~\cite{coleman_reducing_2014}, the principal motion planning library for ROS2, utilizes the Open Motion Planning Library (OMPL)~\cite{sucan_open_2012}. While OMPL yields global solutions, its efficacy in real-time and dynamic applications, executed through \textit{MoveIt Servo}\footnote{\href{https://moveit.picknik.ai/main/doc/examples/realtime\_servo/realtime\_servo\_tutorial.html}{https://moveit.picknik.ai/main/doc/examples/realtime\_servo/-realtime\_servo\_tutorial.html}}, frequently results in singularities, necessitating robot movement away from objectives for recovery. We instead chose Riemannian Motion Policy (RMP)~\cite{cheng_rmpflow_2020}, which is distinctively engineered for dynamic environments and supports real-time operation and collision avoidance.

Although RMP requires extensive fine-tuning, recent Nvidia Isaac Sim updates have significantly streamlined this process by introducing a graphical user interface application. The decision to employ this algorithm was based on its high adaptability for real-time planning in dynamic environments, which is a crucial requirement for an efficient teleoperation system. Furthermore, Nvidia provided example files for UR10 robots, which we successfully adapted for the UR3 and Baxter robotic systems.

\subsection{Robots and End-effectors}\label{sec:gripper}

\begin{figure}[t]
  \centering
  \includegraphics[width=\linewidth]{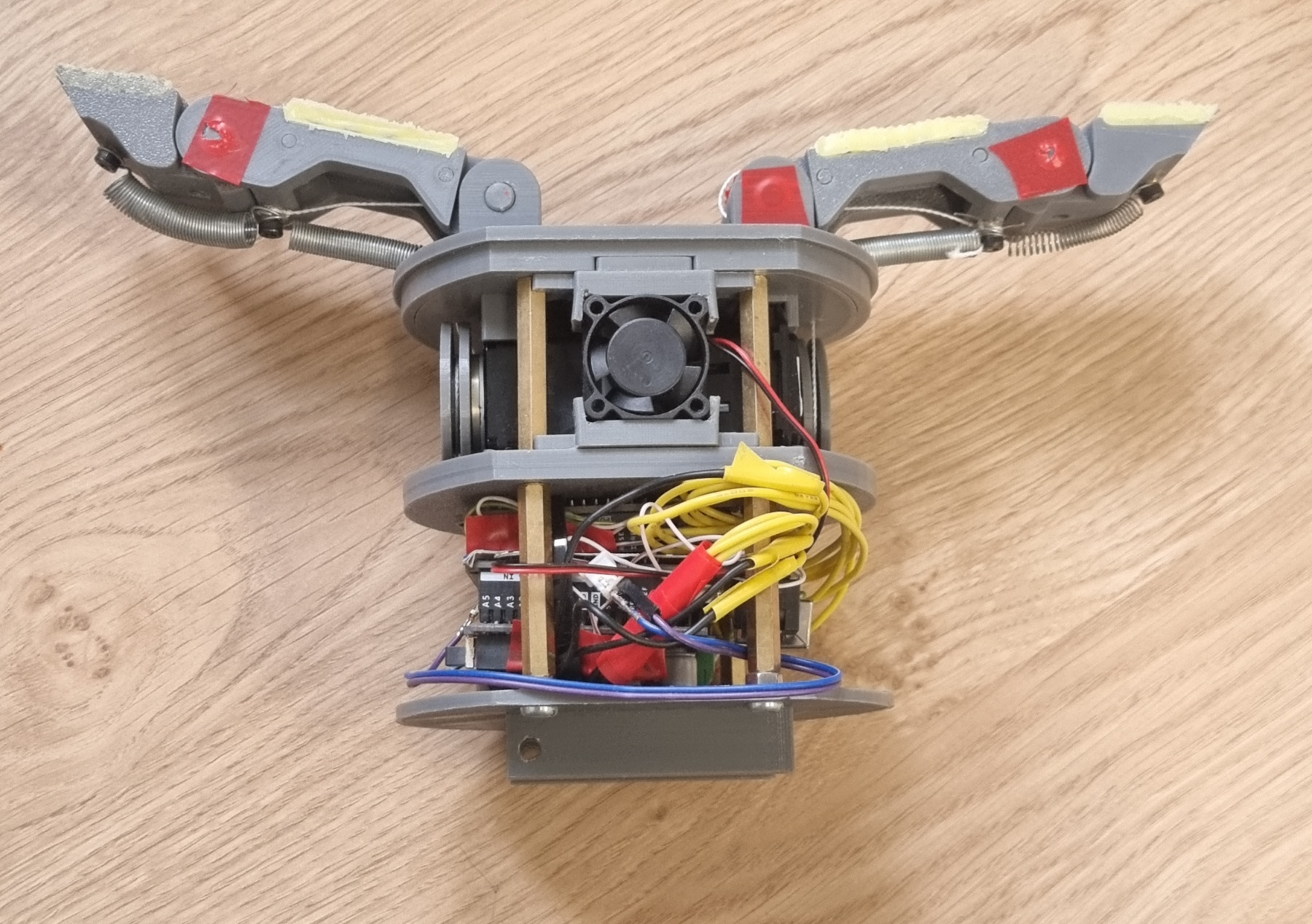}
  \caption{Photo of the modified T42 Gripper with the second level added for the control boards}
  \label{fig:gripper}
\end{figure}

For our experiment we used 3 different grippers, Baxter robot default linear gripper, depicted in Figure \ref{fig:overview}, used with the Baxter robot, a modified Yale T42 gripper~\cite{noauthor_yale_2023}, shown in Figure \ref{fig:gripper}, used with the UR3 robot and a Robotiq 2F-140 gripper ~\cite{noauthor_robotiq_2024}, shown in Figure \ref{fig:experimental_setup_japan}, used with the UR5e robot. The difference in gripper was informed due to our belief that hardware played a part in teleoperation performance. These 3 grippers have varying heights and opening lengths. The Baxter gripper was chosen as a pure linear gripper that is fully integrated with the Baxter robot. The Yale T42 gripper is a 2-finger gripper designed with individual finger motion, causing a significant difference in height between the opened and closed state. Finally, the Robotiq is a blend of a liner gripper and a finger gripper, giving it a larger opening between the 2 fingers at the cost of a change in height between its 2 states.

The Baxter gripper is controlled using the same control method as Baxter, which is a ROS2 control package, communicating with the real robot using a ROS2 to ROS1 bidirectional bridge using the joints states provided by Isaac Sim following the setup described in Section \ref{sec:telesimdetails}. This package was adapted from the ROS1 package provided by Rethink Robotics to ROS2\footnote{\url{https://github.com/CentraleNantesRobotics/baxter\_common\_ros2}} and modified to allow more control of the gripper. The change of gripper state is controlled by a custom ROS2 node listening to the state of the VR controller, as described in Section \ref{sec:controllers}, and publishes a trigger to the digital clone in Isaac Sim whenever the trigger button of the controller is pressed. Thus, this causes the digital clone joint states to be updated and forwarded to the real robot using the ROS2 control package.

The Yale T42 gripper was also chosen due to its 3D printable components and individually controllable fingers, each utilizing two Dynamixel motors. This configuration aligned with the requirement of mapping individual finger movements to the Senseglove interface. An interface was developed to address the SDK's lack of ROS2 integration. This involved developing a ROS2 package to incorporate the SDK into a ROS2 control interface to enable seamless integration with TELESIM.

Initial testing revealed limitations in the control board's communication capabilities. The single high-speed UART channel resulted in significant latency between issuing a command and its execution. Furthermore, the intended implementation of haptic feedback from motor data to the Senseglove required a higher data transmission rate than the virtual UART communication could support. A dual Arduino board setup employing I2C communication was implemented to address these issues, effectively resolving the bandwidth constraints. The decision was made to forgo haptic feedback implementation for this paper, as it would have been the sole control method featuring this capability, potentially complicating comparisons between control methods in user surveys and experiments. Although vibration on the Vive controller could theoretically serve as a form of haptic feedback, this approach would introduce additional variables requiring evaluation, such as vibration strength and delay, to assess their impact on user performance and perception.

We had to alter the original gripper's mechanical design by incorporating two Arduino boards in the gripper. A secondary level was introduced between the motor and the robot attachment point to accommodate the control boards. This change is shown in Fig. \ref{fig:gripper}. This solution was adopted after attempts to mount the boards on the gripper's side, resulting in cable entanglement and damage during robot operation due to spatial constraints. While the additional level resolved the cable management issues, it extended the gripper's length by 5 cm, consequently reducing the UR3 robot's already limited workspace.

The finger control mechanism uses a fishing line under tension driven by the motor. Initial tests revealed that the original fishing line lacked the necessary strength for teleoperation tasks, frequently breaking during use. Although stronger fishing lines were considered, they imposed excessive strain on the motors. To preserve motor integrity, the decision was made to retain the weaker fishing line despite its limitations. The final modification involved the removal of the final degree of freedom on each finger. This alteration was implemented as the original finger configuration hindered the grasping of cubic objects without providing significant advantages for the objectives of this paper. The modifications to the Yale T42 gripper, while addressing critical functional requirements, introduced trade-offs in terms of workspace reduction and grasping force limitations. These adaptations were necessary to meet the specific demands of the teleoperation experiment within the constraints of the available hardware used in our experiments in Sec. \ref{sec:experiments}.

The gripper is controlled using a custom ROS2 package that listens to the joint states given by the digital clone in Isaac Sim and sends them through I2C to the first Arduino board. It republishes the states of the motor if needed to provide an optional force feedback mechanism to the Senseglove. This package runs in addition to a ROS2 control plugin provided by the Universal Robot company\footnote{\url{https://github.com/UniversalRobots/Universal\_Robots\_ROS2\_Driver}}, needed to control both the UR3 and the UR5e.

Finally, the Robotiq 2F-140 gripper ~\cite{noauthor_robotiq_2024} represents an intermediate design between Baxter's linear gripper, which maintains a constant height in both open and closed states, and the T42 gripper, which exhibits an 11 cm height differential between these states. This height variation introduces an additional variable, as the user needs to estimate the appropriate hand height for successful object grasping. This estimation is simplified in the absence of height changes, as the user can visually determine whether the target object falls within the gripper's reach. Once the gripper state in the digital clone is updated after a VR controller trigger press, a custom node transmits specific commands to the gripper using serial communication.

\section{Experimental Material and Methodology}\label{sec:exp_method}\label{sec:experiments}

\begin{figure*}[t]
\centering
\includegraphics[width=0.80\textwidth]{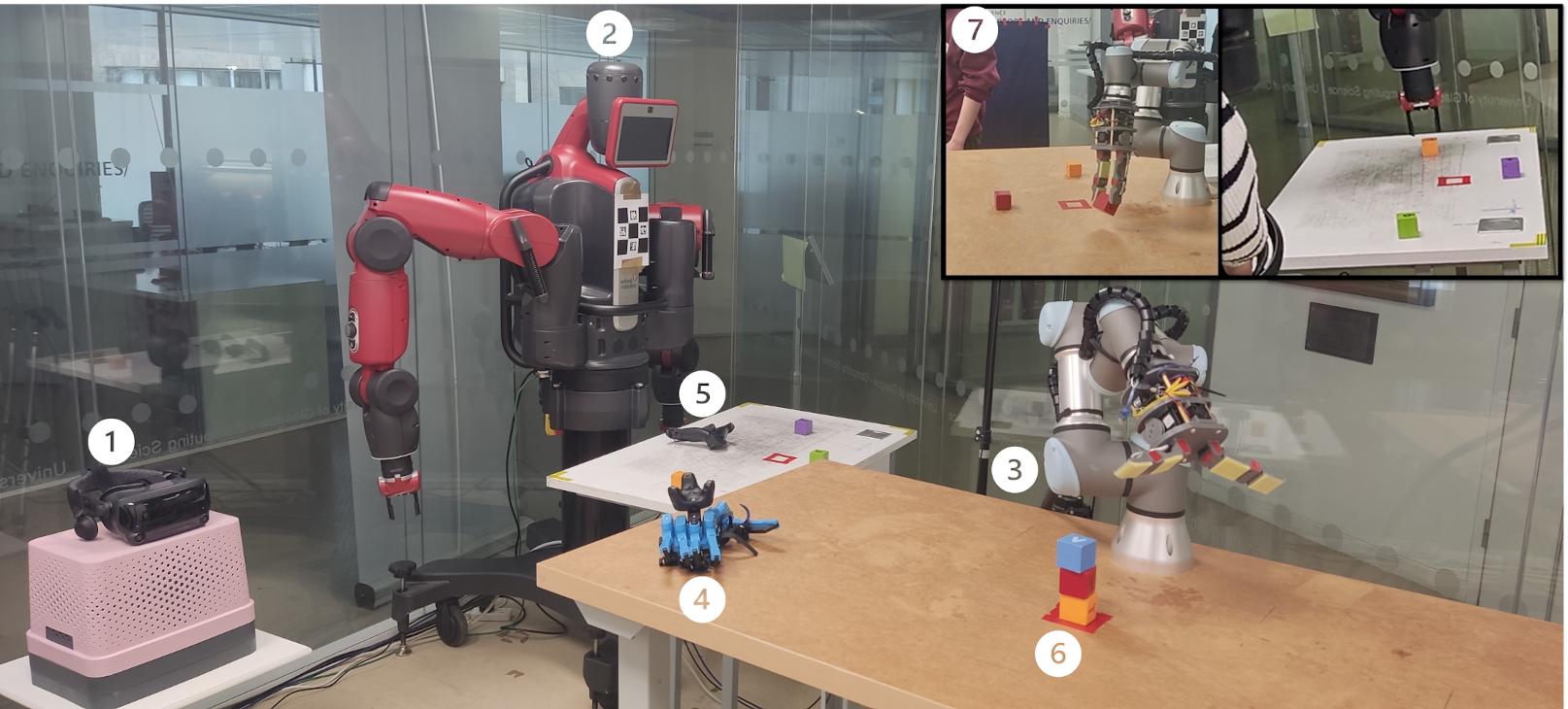}
\caption{Overview of the experimental setup. The Steam Index VR Headset~\cite{noauthor_valve_nodate} is marked as (1) on the far left, which acts as the world's origin. The Baxter robot on the left (2) is controlled by the Steam Index controller (5). In front of it, the UR3 is on the right (3), with the Yale OpenHand T42 gripper~\cite{noauthor_yale_2023}, controlled by the Senseglove and HTC Vive tracker (4) on the left side of the brown table. Additionally, in the upper right corner (7), a view of the starting setup of the task, which consists of 3 cubes in a triangle pattern (described in Section \ref{sec:experimental_setup}), while on the brown table, the cubes are arranged in the goal configuration (6).}
\label{fig:uk_overview}
\end{figure*}

In our experimental setup used in the United Kingdom~\cite{audonnet_telesim_2024}, each robot was positioned in front of a table with cubes arranged in an isosceles triangular pattern, as illustrated in Figure \ref{fig:uk_overview}. The experimental protocol required participants to teleoperate a robot while standing with their backs to the VR headset, which served as both the world's origin and a reference point for the user. The user was tasked with moving and staking the 3 cubes from their starting position to a central position marked with red tape, as visible in Figure \ref{fig:overview}. The bottom 3 images show the evolution of the manipulation from 1 cube in motion to the final completed tower in the bottom right.

Participants were required to complete multiple questionnaires throughout the experiment. Initially, they completed a brief demographics questionnaire, which included age, gender, and experience with virtual reality, robots, and wearable gaming devices (e.g., Wiimote). These questions utilized a 5-point Likert scale to assess each participant's perception and understanding. After teleoperating a robot, users answered two additional questionnaires. The first was the Single Ease Question (SEQ) ~\cite{hodrien_review_2021}, regarded as an effective end-of-task metric. The second was a raw NASA-TLX, described in Section \ref{sec:background}, employing a 7-point Likert scale instead of the 21-point scale, as research has shown that the 21-point scale does not enhance questionnaire reliability ~\cite{ruan_comparing_2017, lewis_user_2017}. This research was validated by the University of Glasgow Ethics Committee (Application Number \textit{300220026})

At the end of the teleoperation experiment, participants completed the Negative Attitude Towards Robots (NARS) questionnaire, described in Section \ref{sec:background}. This study exclusively employed elements from the S1 subscale, which assesses negative attitudes towards situations involving robot interactions (cf. Section \ref{sec:background}). The other subscales were deemed irrelevant as they addressed hypothetical future scenarios with robots or general interactions and conversations with robots. It is worth noting that the use of a partial NARS survey is not unprecedented, as Naneva \textit{et al.} ~\cite{naneva_systematic_2020} reported that the NARS-S2 scale was the most widely used in their systematic review of attitudes. This approach also enables direct comparison with the findings of Bartneck \textit{et al.}~\cite{bartneck_influence_2007}, who utilized NARS to assess trust across seven different countries and are, as far as we are aware, the only researchers to have compared the Trust towards robots for teleoperation of the robotic arm across countries.

To explore the impact of the capabilities and limitations of the robotic hardware used for teleoperation, as well as the impact of demographics and previous experience, we deployed TELESIM on three different robots: the Rethink Robotics Baxter robot, the UR3 and the UR5e. Baxter and UR3 were used to conduct our experiments at the University of Glasgow in the UK. At the same time, the Universal Robot 5e was at the National Institute of Advanced Industrial Science and Technology (AIST) Waterfront Center in Tokyo, Japan. This section explores the difference in setup and environment from the user survey developed in the UK. Due to time constraints, we were required to recruit our participants through an agency. Thus, we had access to a more significant reach than in the UK, reaching participants who were not involved in academia (i.e. undergraduate and postgraduate students and academics). Additionally, participants were financially compensated for participating in the user survey, unlike in the UK, where they participated for no monetary rewards. In addition, this research was internally approved by the Ergonomics Experiment Committee of the Life Sciences Experiment Management Office in the Environmental Safety Department of AIST with the following application number: \textit{2023-1384}. To keep our analysis consistent, we recruited participants within the same age range as described below:

\begin{itemize}
    \item Japan--age range of 25.32 $\pm$ 6.26; and 28 male and 9 female;
    \item The United Kingdom--range age of 27.81 $\pm$ 7.93 and 29 male and 8 female.
\end{itemize}

\subsection{Robotic Setups in the UK and Japan}\label{sec:experimental_setup}

A comprehensive analysis of their strengths and weaknesses is necessary to help us better understand the three robots employed in this study. Table \ref{tab:robot_stats} summarizes the robots' capabilities.

Baxter exhibits the longest reach at 1210 mm, followed by the UR5e with 850 mm and the UR3 with 350 mm. However, Baxter's extended reach introduces end effector position variability ~\cite{cremer_performance_2016}, a characteristic not observed in the Universal Robots. This variability is denoted as "no stability" in Table \ref{tab:robot_stats}.

The primary distinctions between the two Universal robots lie in the UR5e's extended reach, gripper configuration, and control methodology. The UR3 was equipped with a modified Yale T42 gripper ~\cite{noauthor_yale_2023}, measuring 21 cm when fully closed. Due to its size relative to the robot's body, restrictive joint limits were implemented to prevent gripper-body collisions, consequently limiting the robot's operational area. Conversely, the UR5e utilized a Robotiq 2F-140 gripper, measuring 23.3 cm when closed. For consistency, identical joint limits were applied to the UR5e; however, its longer reach mitigated the impact of these restrictions.

A notable difference between the grippers is the height differential between their open and closed states, as mentioned in Section \ref{sec:gripper}. The Robotiq gripper exhibits a 2.35 cm difference, while the T42 demonstrates an 11 cm difference. This disparity increased users' difficulty in estimating the appropriate gripper height for cube grasping with the UR3, as explained in Section \ref{sec:experimental_setup}.

The T42 gripper's design introduced additional grasping challenges. As reported in TELESIM ~\cite{audonnet_telesim_2024}, the Yale gripper's limited closing force occasionally resulted in cube slippage during transit. This limitation was not observed with the Robotiq gripper.
Lastly, the robots' control methods differed. The UR3 was controlled via a Senseglove with an attached Vive tracker, while both the UR5e and Baxter utilized a Valve VR controller.

Distinct control methods were employed for each robot. The Baxter robot was operated using a Steam Index VR controller, as depicted in Figure \ref{fig:uk_overview} (5). In contrast, the UR3 robot was controlled via a Senseglove development kit, enabling the mapping of individual finger movements, with an HTC Vive Tracker mounted atop the hand (4). The control of a modified T42 gripper from the Yale OpenHand project~\cite{noauthor_yale_2023}, attached to the UR3, was limited to the user's thumb and index finger movements (3) in Figure \ref{fig:uk_overview}.

For our Japan experiments, the TELESIM framework underwent minimal modifications, primarily involving the URDF file, which now loads a UR5e robot and replaces the T42 Gripper with a Robotiq 2F-140 gripper~\cite{noauthor_robotiq_2024}. This change was required given the significant design flaws of the T42 gripper, as discussed in Section \ref{sec:gripper}. The Digital Twin was updated to reflect the new location and incorporate revised virtual safety settings, constraining the motion planner from approaching certain areas due to safety considerations. Owing to the compatibility between Universal Robots 3 and 5e, we were able to utilize the same ROS2 controller developed during the initial TELESIM implementation.

\begin{figure}[t]
    \centering
    \includegraphics[width=0.8\linewidth]{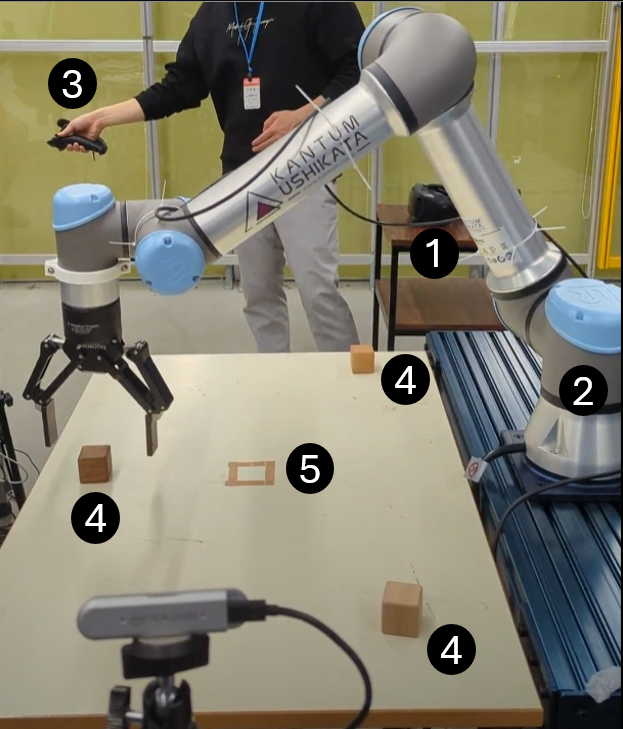}
    \caption{Overview of our experimental setup in AIST. The Steam VR headset (1) acts as the world's origin, as seen behind the robot and the user. The UR5e robot (2) in front is controlled by a Steam Index VR controller (3). There are 3 cubes set up in their starting position (4) in a triangular pattern similar to ~\cite{audonnet_telesim_2024}. The empty square in the middle of the table (5) represents the location where the user should stack all the cubes, resulting in a tower of 3 cubes.}
    \label{fig:experimental_setup_japan}
\end{figure}

Consistent with our previous UR3 experiment, the gripper's collision parameters were configured based on its closed state to prevent hardware damage. To maintain analytical consistency, the joint position and speed limits applied to the UR5e were identical to those used in the UR3 experiment despite the UR5e's extended reach. Due to the extended reach, our previous cube placement was no longer suitable, as the robot was able to reach all the locations on our available table. Nevertheless, the experimental design aimed to preserve task difficulty by positioning cubes in a manner that required participants to adjust their body position to reach all cubes. Depending on the target cube, this typically involves a single step in various directions. The cube arrangement and the step required to access the leftmost cube are depicted in Figure \ref{fig:experimental_setup_japan}.


\begin{table}[t]
    \caption{Robots Specifications}
    \centering
    \renewcommand{\arraystretch}{1.5}
    \begin{tabular}{c|c|c|c|c}
        \multirow{2}{0.12\linewidth}{\centering\textbf{ Robot Type}}& \multirow{2}{0.18\linewidth}{\centering\textbf{ Control Method}}& \multirow{2}{*}{\centering\textbf{ Gripper}}&
        \multirow{2}{*}{\centering\textbf{ Stable}}& 
        \multirow{2}{0.14\linewidth}{\centering\textbf{ Reach (mm)}} \\
        &&&&\\
        \noalign{\hrule height 1.5pt}
        Baxter & VR Controller & Rethink Gripper & No & 1210 \\
        \multirow{2}{*}{\centering UR3} & 
        \multirow{2}{*}{\centering Senseglove} & 
        \multirow{2}{0.2\linewidth}{\centering Modified Yale T42 Gripper} & 
        \multirow{2}{*}{\centering Yes} & 
        \multirow{2}{*}{\centering 350} \\
        &&&&\\
        UR5e & VR Controller& Robotiq 2F-185 & Yes & 850\\
    \end{tabular}
    \label{tab:robot_stats}
\end{table}

\section{Evaluation}\label{sec:eval}

\subsection{Experimental Results} \label{sec:results}
\begin{figure}[t]
  \centering
  \includegraphics[width=\linewidth]{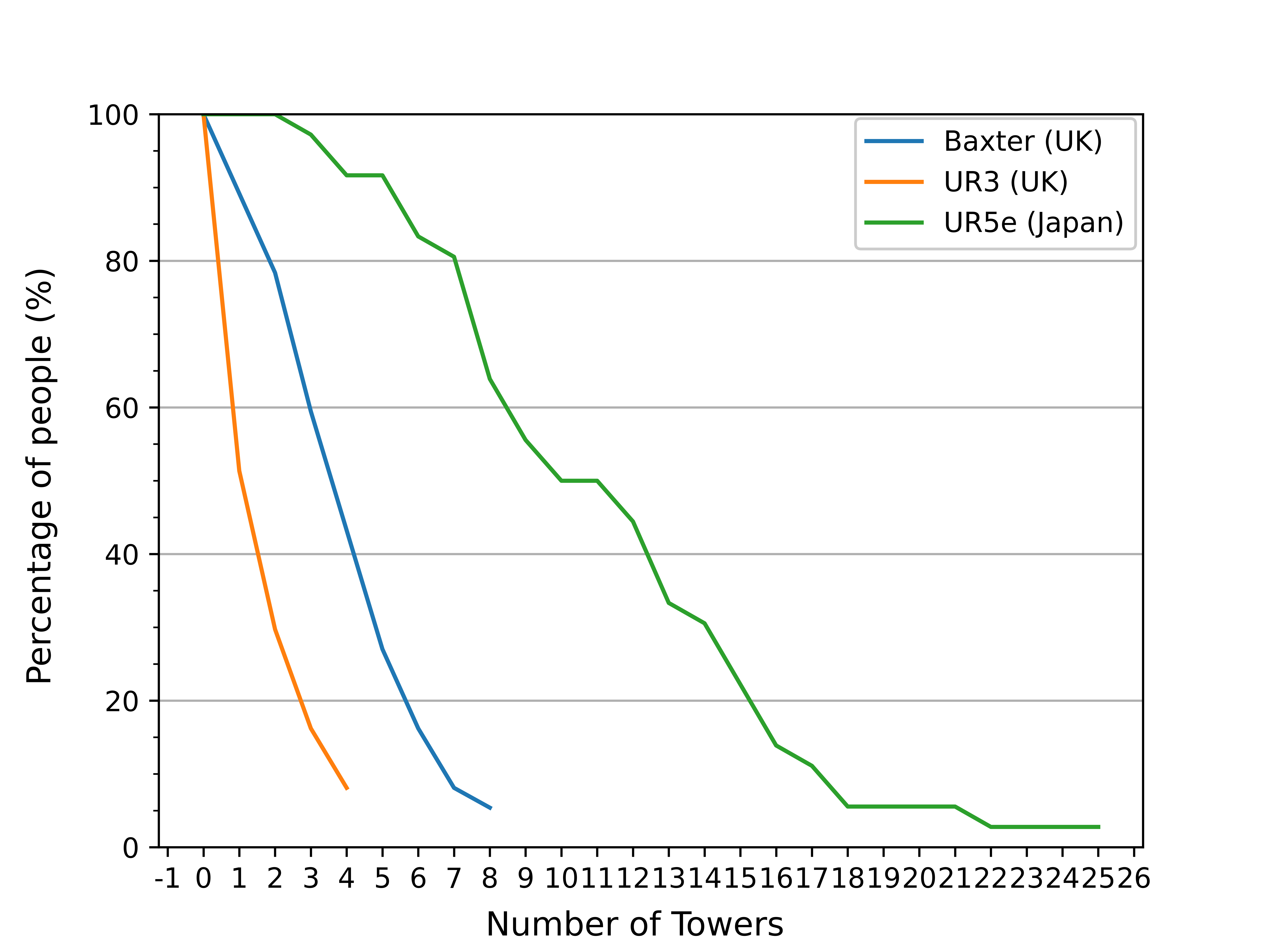}
  \caption{Population percentage for each tower for Baxter in blue, the UR3 in orange, and the UR5e in green}
  \label{fig:performance}
\end{figure}

\begin{figure}[t]
  \centering
  \includegraphics[width=\linewidth]{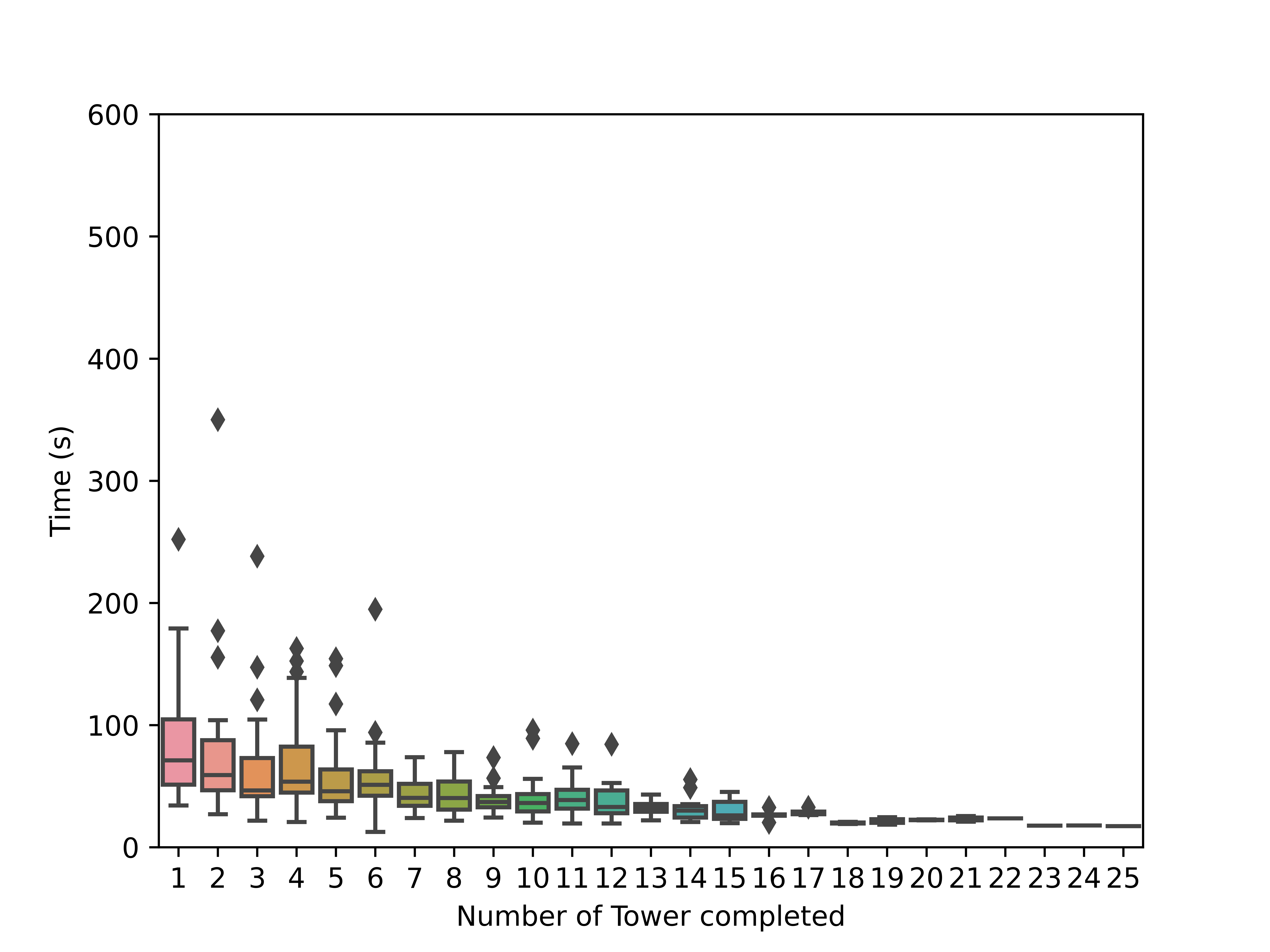}
  \caption{Average Time Taken for each tower completed for the UR5e robot.}
  \label{fig:aist_tower}
\end{figure}
\begin{figure}[t]
  \centering
  \includegraphics[width=0.85\linewidth]{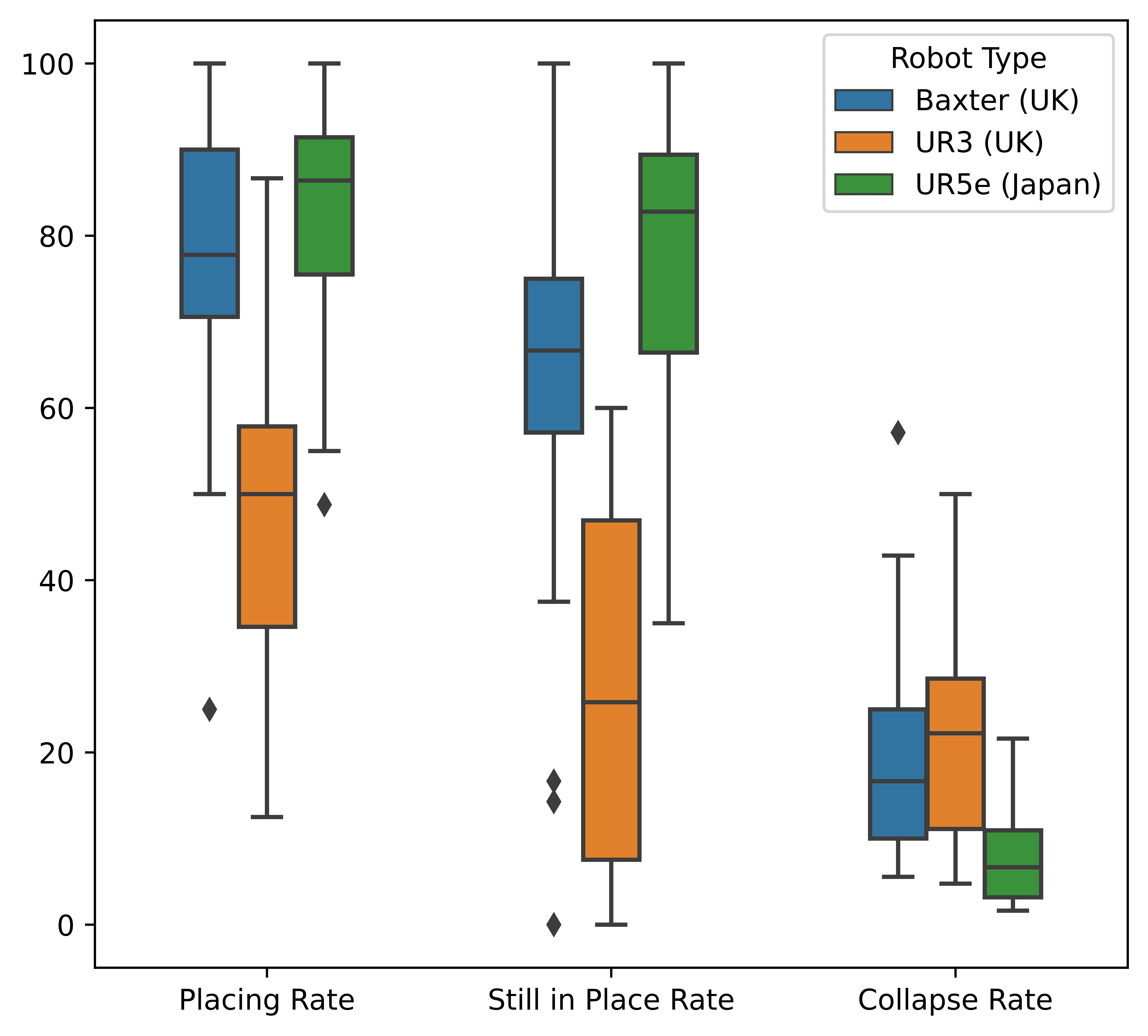}
  \caption{Ratio of different statistics collected during the experiment. The Placing Rate is calculated as the number of place actions over the number of picking actions. The Collapse Rate is calculated as the number of Collapse actions over the number of picking actions. The Still in Place Rate is calculated as the number of Place actions minus the number of collapses over the number of picking actions, effectively rating the tower's stability.}
  \label{fig:stats}
\end{figure}
\begin{figure*}[t]
  \centering
  \includegraphics[width=\columnwidth]{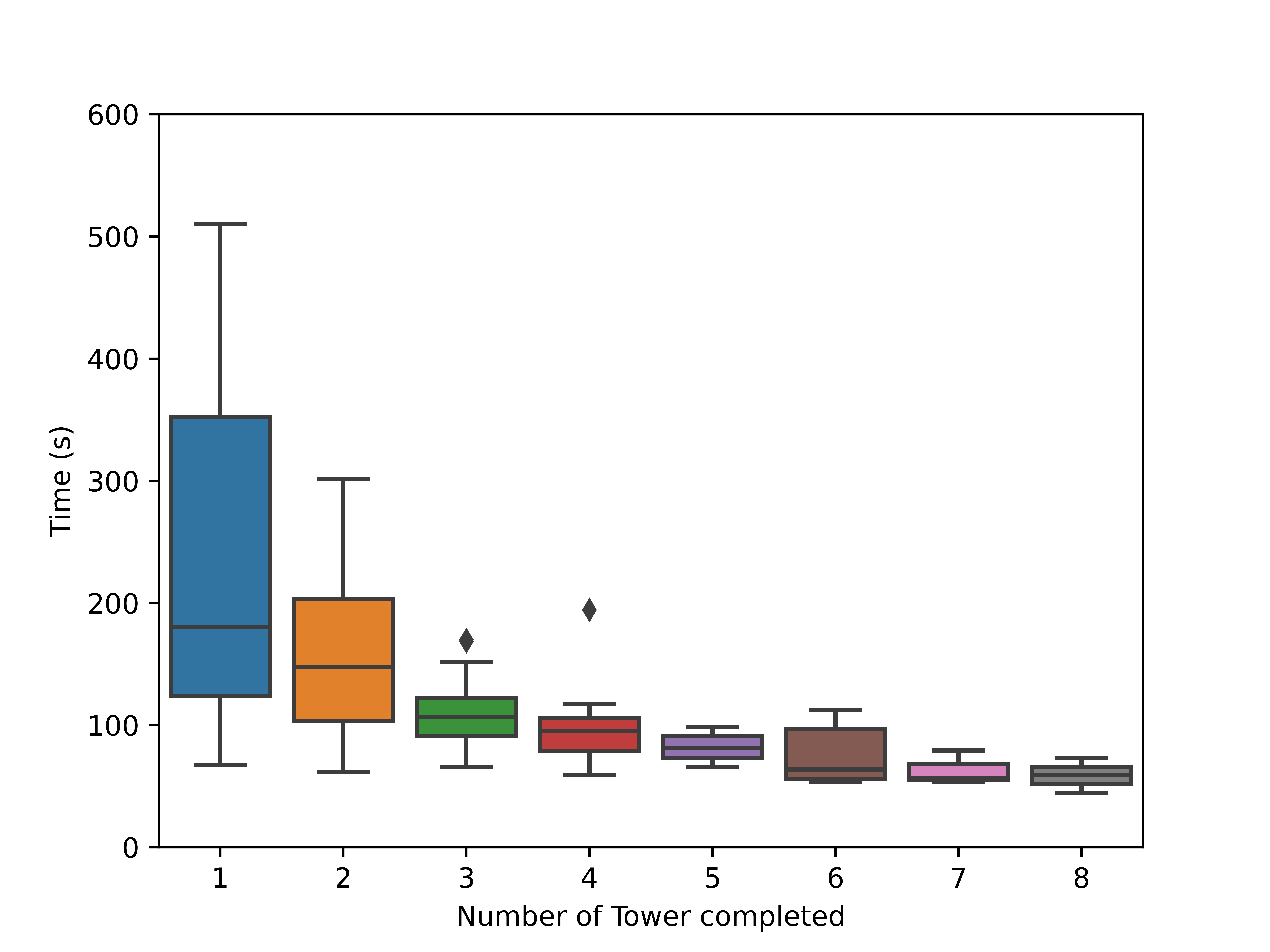}
  \includegraphics[width=\columnwidth]{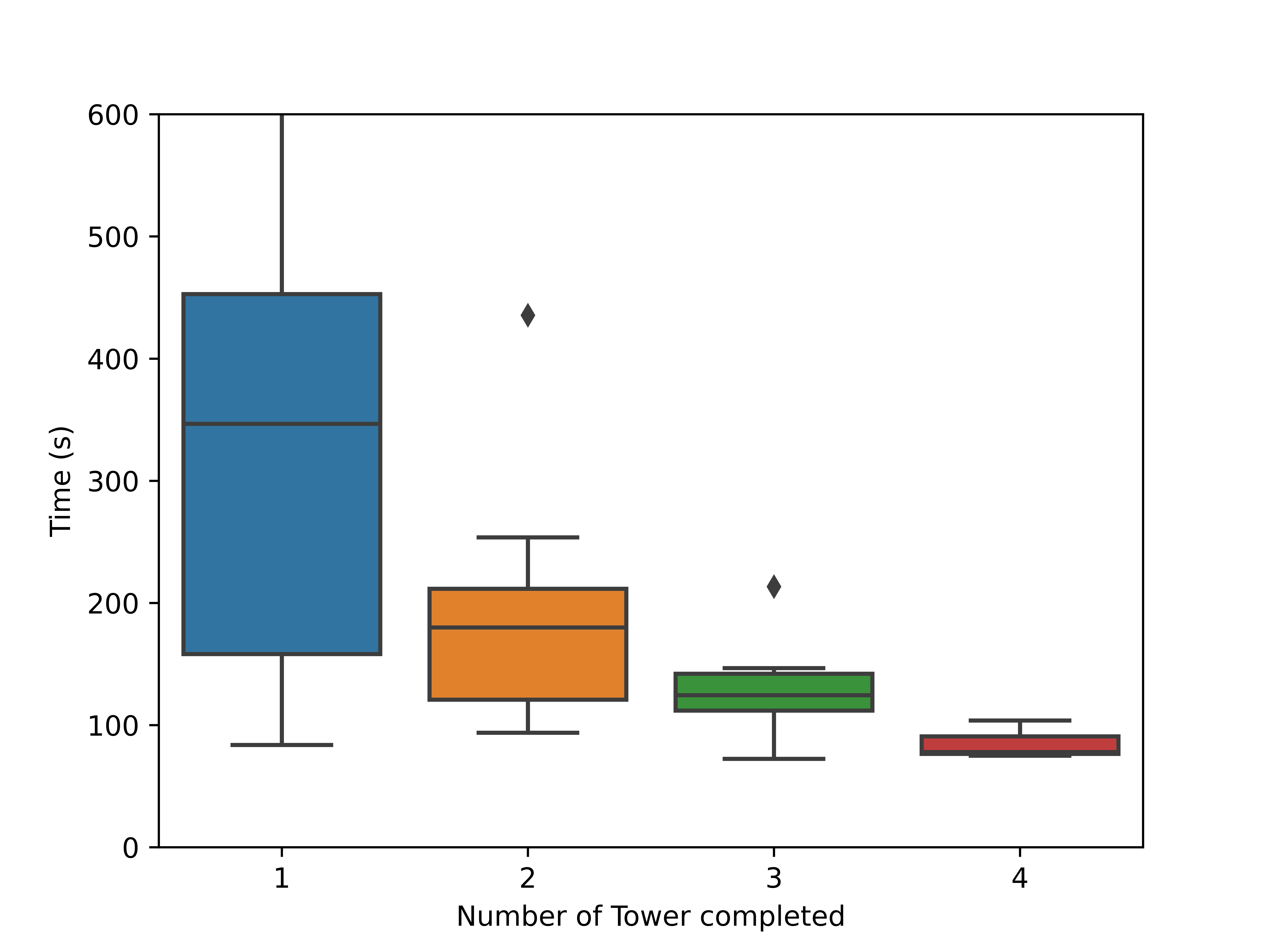}
  \caption{Average Time Taken for each tower completed for the Baxter robot (left) and the UR3 robot (left).}
  \label{fig:baxter_tower}
\end{figure*}


Figure \ref{fig:performance} and Figure \ref{fig:aist_tower} illustrate that all participants using the UR5e successfully constructed a minimum of 3 towers within the 10-minute timeframe, with an average construction time of less than 100 seconds per tower. Moreover, subsequent towers were completed in progressively shorter duration. One participant notably constructed 25 towers, averaging 40 seconds per tower. The mean number of towers constructed across all UR5e users was 10.50.

In comparison, the average number of towers constructed using the Baxter robot was 3.25, while the UR3 yielded an average of 1.03 towers. This performance disparity is shown in Figure \ref{fig:performance}. Furthermore, for both the Baxter and UR3 robots, only participants who constructed 4 or more towers achieved completion times under 100 seconds, as shown in Figure \ref{fig:baxter_tower}.

Figure \ref{fig:performance} depicts a steep performance improvement for the UR5e up to the 18th tower, followed by a plateau for the remaining seven towers. This plateau is attributed to the fact that only two participants managed to build 21 and 25 towers, respectively. The exceptional performance of these participants suggests prior experience with robots or robotic teleoperation. However, the current questionnaire design, being self-reported, may not accurately capture this experience due to potential over- or under-estimation of skills by participants.

An informal discussion with the participant who constructed 25 towers revealed professional experience with robots. This background appears to have provided spatial skills that are directly applicable to robot teleoperation tasks.

Additional statistics were collected to investigate the relationship between teleoperation performance and robot capabilities, as shown in Figure \ref{fig:stats}. It is important to note that the teleoperation user surveys for Baxter and UR3 were conducted in the UK, while those for the UR5e were performed in Japan. The differences in experimental setup are detailed in Section \ref{sec:experiments}.

The \textit{Placing Rate}, defined as the ratio of successfully placed cubes to the total number of cube picks, aligns with the tower construction results. The UR5e demonstrated the highest values, followed by Baxter and the UR3. Interestingly, the difference between Baxter and the UR5e was not statistically significant (P $>$ 0.1). However, the UR3 exhibited a statistically significant difference (P $<$ 0.001) compared to both Baxter and the UR5e. This disparity can be attributed to the advantages and disadvantages outlined in Section \ref{sec:exp_method} and summarized in Table \ref{tab:robot_stats}, with gripper design and control methods being the primary differentiating factors.

The \textit{Still in Place} rate, representing the frequency of cubes remaining in their correct position without collapse, revealed a statistically significant advantage for the UR5e over Baxter (P $<$ 0.01). This allows rejecting the null hypothesis that users achieved similar tower stability with Baxter and the UR5e. This conclusion is further supported by the difference in collapse rates, with the UR5e showing a statistically significant difference compared to Baxter (P $<$ 0.1).

The UR5e and UR3 exhibited statistically different collapse rates, with the UR5e performing better than the UR3 and Baxter. While the UR5e outperformed both robots, its advantage over Baxter primarily stemmed from better stability. Conversely, the UR3's higher collapse rate can be attributed to other factors, particularly its control method, as noted in Section \ref{sec:exp_method}.

\begin{figure}[t]
  \centering
  \includegraphics[width=0.85\linewidth]{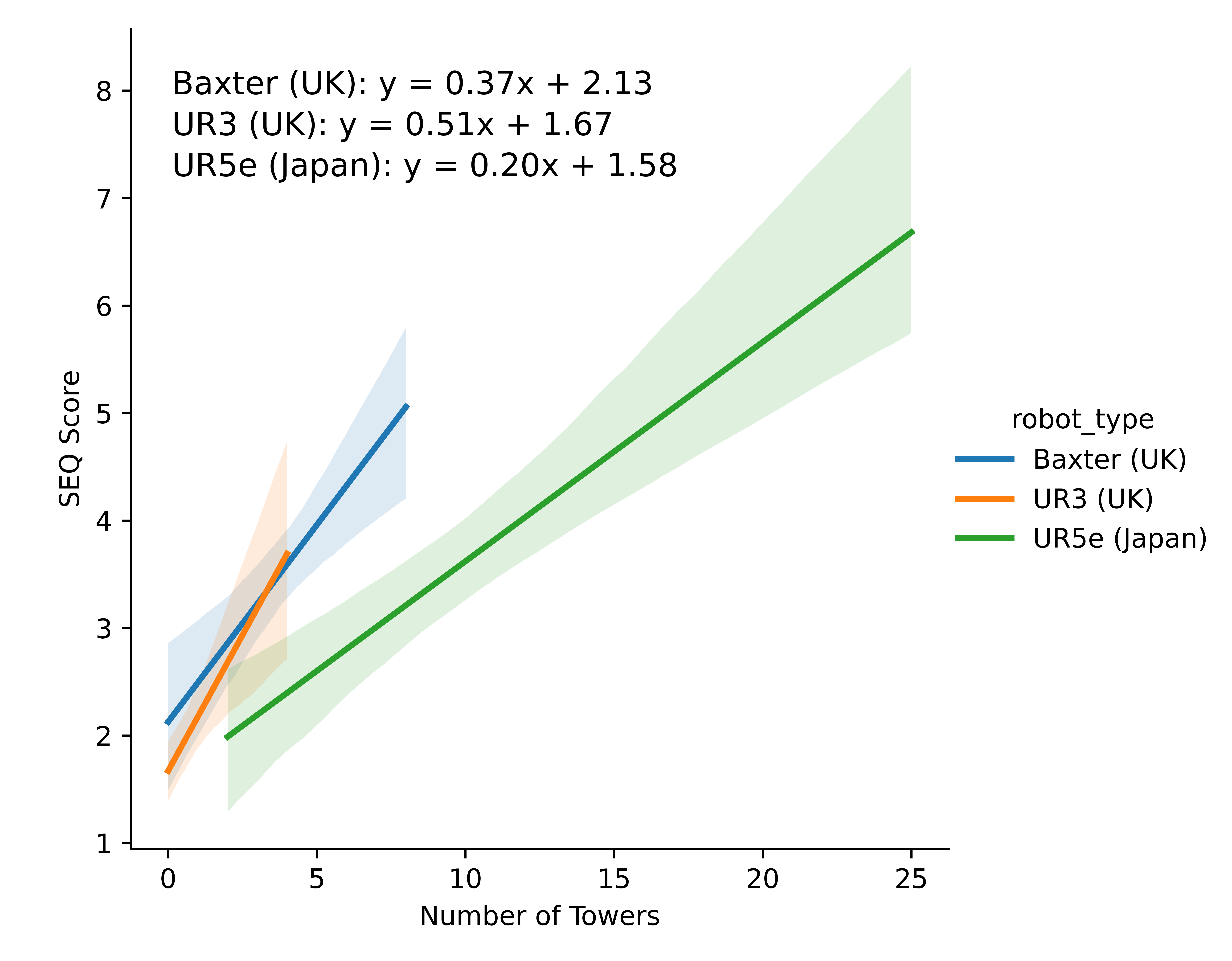}
  \caption{Regression plot showing the relationship between SEQ and the number of towers for each individual robot. }
  \label{fig:seq}
\end{figure}

Interestingly, the results of the Single Ease Question (SEQ) were statistically indistinguishable for Baxter and the UR5e (P $>$ 0.01) despite the disparity in tower construction performance. Baxter achieved a mean SEQ score of 3.32 (SD = 1.27), the UR3 scored 2.19 (SD = 1.14), and the UR5e scored 3.72 (SD = 1.54). Higher scores indicate greater perceived accessibility. We hypothesize that the UR5e's enhanced stability may have led users to perceive a higher tower construction potential than Baxter. Notably, participants were not informed of the maximum achievable tower count before completing the questionnaire. It appears that users estimated the theoretical maximum based on their experience, resulting in comparable SEQ scores for both robots despite performance differences.
As expected, there is a clear relationship between the number of towers and a higher SEQ score, as can be seen in Figure \ref{fig:seq}

Our analysis primarily focused on the robots' capabilities. However, it is important to note that the UR5e experiment was conducted in Japan, and we hypothesize that demographic factors may have influenced the results. The decision to conduct this user survey in Japan was based on research by Nam et al. ~\cite{nam_trust_2018}, which suggests that user workload varies depending on the level of trust in a system. Bartneck et al. ~\cite{bartneck_influence_2007} indicate that participants from different countries exhibit varying levels of trust toward robots.

The Japanese participants self-reported higher levels of experience with robots (mean 3.86 ± 1.27) compared to the British participants (mean 2.76 ± 1.24). This difference is statistically significant (P $<$ 0.001) and can be attributed to the greater exposure of the Japanese participants to robots in their daily lives. Interestingly, the Japanese participants reported less experience with wearable technology (mean 2.81 ± 1.47) compared to the British participants (3.86 ± 0.87), a difference that is also statistically significant (P $<$ 0.01). However, virtual reality (VR) experience was statistically similar between the Japanese participants (mean 3.25 ± 1.38) and the British participants (mean 2.97 ± 1.20) participants. These findings suggest that previous experience with robots and wearables varies depending on the user's background. In contrast, VR experience remains consistent across both demographics, possibly due to the recent widespread adoption of VR technology ~\cite{hamad_how_2022}.

It is important to note that we are unable to directly correlate previous experience with robots or wearables to teleoperation performance due to the differences in the systems users operate. However, we were able to exclude previous VR experience as a contributing factor to performance differences.

\subsection{Operator Workload}\label{sec:nasa-results}

\begin{figure}[t]
  \centering
  \includegraphics[width=0.85\linewidth]{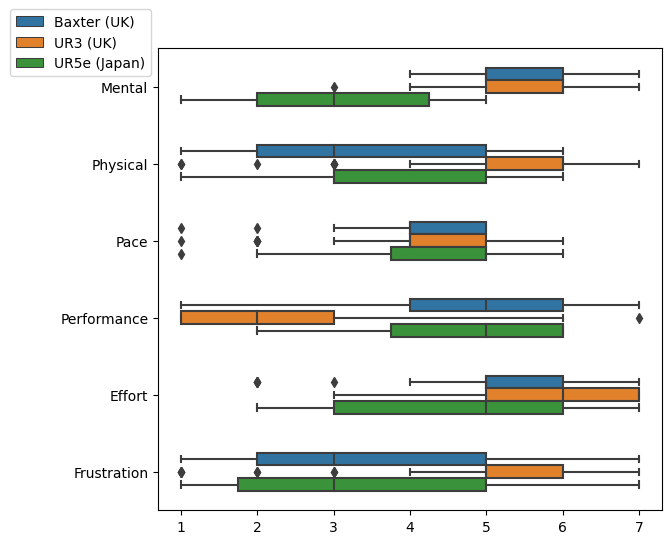}
  \caption{Boxplot representing individual results from the NASA task survey, in which a lower score indicates lower difficulty. Blue represents the experiment carried out with the UR5e in Japan. Orange represents the experiment with the UR3 in the UK. Green represents the experiment with Baxter in the UK.}
  \label{fig:nasa_boxplot}
\end{figure}

Figure \ref{fig:nasa_boxplot} illustrates the individual results for each subscale or dimension of the NASA-TLX survey. The subscales are described as follows:
\begin{itemize}
\item Mental: This subscale examines the mental workload imposed on the user during task performance. The UR5e demonstrates the lowest mental workload, statistically significant at the 95\% level compared to Baxter and the UR3, which show similar results. Interestingly, the control method difference between Baxter (VR controller) and UR3 (Sensglove) does not impact mental workload, nor does the UR3's weaker gripper. The UR5e's stability and range likely explain this difference, providing users with greater freedom of movement without concern for robotic constraints.
\item Physical: This subscale assesses the user's physical workload during task execution. The UR3 exhibits a statistically significant difference compared to the other two robots, attributable to its distinct control method. The UR3, controlled by the Senseglove, requires users to maintain a flat, extended hand position, unlike the VR controller, which allows for a more relaxed hand posture closer to the body.
\item Pace: This subscale examines the user's perception of time pressure. The results are statistically consistent across all robots, which is expected since all experiments were conducted under uniform time constraints.
\item Performance: This subscale examines the user's perceived task performance. The UR3 demonstrates the lowest perceived performance, statistically significant at the 95\% level compared to the other two robots. This aligns with the Single Ease Question results described in Section \ref{sec:results}, as both metrics evaluate task difficulty.
\item Effort: This subscale investigates the perceived effort spent by the user to complete the task. The UR3 shows the highest effort expenditure, which is statistically significant at 95\% compared to Baxter and the UR5e. The consistently high effort scores across all robots corroborate the assertion that robot teleoperation is challenging for non-experts, as noted in previous research ~\cite{rea_still_2022, muto_touch_2012}.
\item Frustration: This subscale explores the user's frustration and stress levels during the experiment. The UR3 leads with a statistically significant difference over Baxter and the UR5e, likely due to the disadvantages outlined in Section \ref{sec:exp_method}, particularly its limited range and weak gripper. Notably, this is the only subscale where participants utilized the full range of possible answers for all robots. This suggests that frustration tolerance is highly individual and may not be an ideal metric for evaluating teleoperation system performance.
\end{itemize}

The above NASA results indicate a relationship between controller type and user impact, particularly in physical and emotional domains, manifesting as frustration and stress. Additionally, robot type influences user mental workload, with the UR5e's combination of large reach and high stability resulting in reduced mental strain compared to robots possessing only one of these qualities, such as Baxter and the UR3.

The hypothesis that demographics play a significant role can be discarded, as additional statistics such as \textit{collapse rate} and \textit{still in place rate} are similar for both Baxter and the UR5e (cf. Figure \ref{fig:stats}). This non-significance is further supported by the similarity across all NASA subscales, except for mental workload.
Finally, these results suggest a trend that, regardless of the robot type, users experience frustration and some degree of physical strain. We postulate that this contributes to the difficulty of teleoperation for non-experts, as noted in previous research ~\cite{rea_still_2022, muto_touch_2012}.

\subsection{Trust Towards Robots}

\begin{figure}[t]
  \centering
  \includegraphics[width=\linewidth]{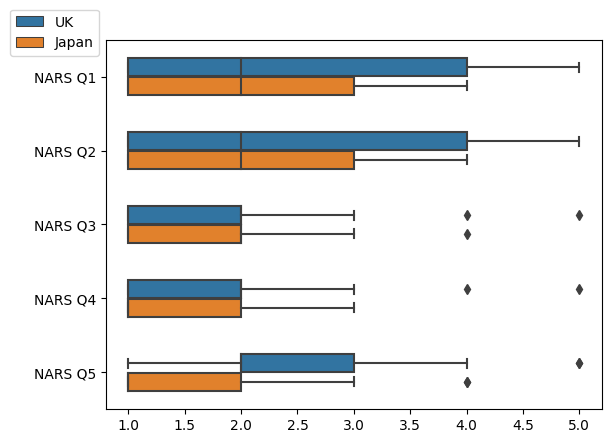}
  \caption{Boxplot representing the total score for the NARS (Negative Attitude Towards Robot) where a low score indicates the user is more trusting of the robot. Blue represents the experiment done in the UK. Orange represents the experiment done in Japan. }
  \label{fig:nars_boxplot}
\end{figure}

Our NARS questionnaire, which evaluates trust, focused solely on elements from the S1 subscales, as described in Section \ref{sec:telesim}. Figure \ref{fig:nars_boxplot} illustrates the individual and combined scores for each question of the S1 subscale. Baxter and UR results are combined, as the questionnaire was administered only once. In retrospect, administering the questionnaire twice—once before the experiments and again after manipulating the first robot—might have provided insights into the progression and impact on trust for each robot.

For completeness, the questions used in our questionnaire, nd in ~\cite{syrdal_negative_2009}, are:

\renewcommand{\labelenumi}{\textbf{NARS Q.\arabic{enumi}}}

\begin{enumerate}[leftmargin=5em]
    \item I would feel uneasy if I was given a job where I had to use robots
    \item I would feel nervous operating a robot in front of other people
    \item I would feel very nervous just standing in front of a robot.
    \item The word "robot" means nothing to me. 
    \item I would hate the idea that robots or artificial intelligence were making judgments about things. 
\end{enumerate}
\renewcommand{\labelenumi}{\arabic{enumi}}

Figure \ref{fig:nars_boxplot} reveals that the Japanese users demonstrated higher trust in robots for questions \textbf{NARS Q.1}, \textbf{NARS Q.2}, and \textbf{NARS Q.5}. However, this difference is statistically significant only for \textbf{NARS Q.5} (P$<$0.05). This general trend results in a statistically significant difference in total scores between the two demographics (P$<$0.05). This disparity may be attributed to the longer exposure of the Japanese participants to robots in their daily lives, as noted by Bartneck et al. ~\cite{bartneck_influence_2007}. Additionally, Japanese culture's lesser distinction between natural and artificial entities, influenced by Buddhist beliefs that do not discriminate against spirits in machines (unlike Christianity), may play a role ~\cite{kaplan_who_2012, bartneck_influence_2007}. Media portrayal of robots may also contribute, with Western media often depicting robots as antagonists. At the same time, Japanese manga presents a more nuanced view where robots can assist in combating human-originated evil ~\cite{bartneck_influence_2007}. The significant difference in \textbf{NARS Q.5} particularly supports this notion, as the concept of robots making decisions often leads to their rebellion in Western media ~\cite{osawa_visions_2022}.

Interestingly, our results appear to contradict the conclusion drawn by Bartneck et al. ~\cite{bartneck_influence_2007} that the Japanese participants are less trusting than the British participants. We hypothesize that this discrepancy may be due to the advancement and proliferation of robots and AI in daily life since the publication of this research work. 

Finally, it is noteworthy that this difference in trust does not appear to affect teleoperation performance, as the Japanese participants achieved the highest number of towers built. This outcome is primarily attributed to hardware differences, given that previous experience with VR-related technology is similar between the Japanese and the British participants, as explained in Section \ref{sec:results}.

\section{Conclusion and Future Work}

This paper presents an extensive exploration of teleoperation's impact on users through three user surveys, encompassing over 70 participants across two countries, utilizing three different robots and two control methods. The study employed NASA-TLX questionnaires to evaluate workload, NARS questionnaires to assess trust toward robots, and SEQ to gauge task difficulty. Additionally, the research investigated trust towards robots across different demographics.

Experimental results demonstrate that participants achieved the highest number of towers built using the UR5e. This superior performance is attributed to the UR5e's enhanced stability and reach, allowing for better cube placement and reduced tower collapse during continued teleoperation. The UR5e also induced the least cognitive stress, while combining the Senseglove and UR3 resulted in the highest physical strain and user frustration (cf. Figure \ref{fig:nasa_boxplot}).

The findings in this paper corroborate previous research ~\cite{rea_still_2022, muto_touch_2012} indicating that teleoperation is challenging for non-experts, characterized by high frustration and physical stress across all robot types. As anticipated, the Japanese participants reported more VR experience than the British participants. Conversely, the Japanese participants reported less experience with wearable technology. The study demonstrated that prior VR experience does not significantly impact teleoperation performance, regardless of the operator's background.

Interestingly, the Japanese participants exhibited higher trust towards robots compared to the British participants, contradicting the conclusions drawn by Bartneck et al. ~\cite{bartneck_influence_2007}. These user surveys also validated the TELESIM framework by enabling international-scale teleoperation across multiple robotic platforms. In addition, the consistent results across demographics proved the user-friendliness of the TELESIM framework, along with its plug-and-play capability, as shown by its deployment on multiple robotic interfaces and end effectors.

While this paper provides an in-depth analysis of teleoperation's impact on users, it focuses solely on direct teleoperation with visual feedback. Other teleoperation methods mentioned in the literature, such as VR headset usage or telepresence robot control ~\cite{dafarra_icub3_2022}, were not explored. Moreover, additional comparison with the state of the art is difficult due to the disparity in the teleoperation task and control methods.

In future work, we aim to use these systems and investigate their impact on teleoperation success and user experience, following the methodology presented in this paper. However, our work on direct teleoperation is not completed. While our experiments seem to indicate that demographics do not play a part in the performance of teleoperation systems, we would need a wider breadth of demographics to show this conclusively. Furthermore, the many differences in hardware during our experiments make it difficult to accurately pinpoint the factors that impact the most, either the teleoperation performance or the user's workload. A deeper investigation of the Senseglove and T42 gripper's performance on the UR3 would enable a more direct comparison with the UR5e and further substantiate our hypothesis that hardware correlates with enhanced performance. This analysis would help elucidate whether the UR3's suboptimal performance stems from gripper limitations, control methodology, or its restricted operational range.

Finally, the exact skill set required to perform teleoperation with minimal training beforehand remains an unsolved problem. The study successfully eliminated prior virtual reality experience as a confounding variable. However, a participant from Japan, who possesses expertise in motion retargeting for robotic teleoperation, demonstrated superior performance by completing the highest number of towers. This observation suggests that spatial abilities or other complex skills may influence task performance.




\bibliographystyle{IEEEtran}
\bibliography{references, lib}

\end{document}